\documentclass{article}
\usepackage{ijcai26}

\usepackage{times}
\usepackage{soul}
\usepackage{url}
\usepackage[hidelinks]{hyperref}
\usepackage[utf8]{inputenc}
\usepackage[small]{caption}
\usepackage{graphicx}
\usepackage{amsmath}
\usepackage{amssymb}
\usepackage{amsthm}
\usepackage{booktabs}
\usepackage{algorithm}
\usepackage{algorithmic}
\usepackage[switch]{lineno}
\usepackage{enumitem}

\title{Distributional Alignment of LLMs and Its Generalization Properties}

\author{
    Nelson Jo
    \affiliations
    Independent Researcher\thanks{Work done as a summer fellow at MIT.}
    \emails
    nelsonjoshuajo@gmail.com
}

\begin{document}

\maketitle

\begingroup
\renewcommand\thefootnote{}
\footnotetext{\hspace*{-1.8em}Accepted at the \emph{IJCAI 2026 Workshop on Augmented Democracies.}}
\addtocounter{footnote}{-1}
\endgroup

\begin{abstract}
Large language models are increasingly used to simulate and probe human behavior, e.g., for surveys. For these applications, a key property is that outputs satisfy \textit{distributional alignment}: to match its output distribution with some reference (in this case, human) distribution. An obstacle in achieving distributional alignment is generalizing from small training data. In this work, we study \textit{(1)} how to distributionally align LLMs; and \textit{(2)} how aligned LLMs might generalize to unseen samples and unrelated topics and tasks. Using two large-scale benchmarks, we find that LLMs can indeed achieve distributional alignment and generalize to unseen questions of the same domain. Cross-topic generalization depends on the relationship between topics (e.g., questions about politics are related to trust in institutions). However, aligned LLMs can degrade in surprising ways (e.g., lose their creative ability), suggesting that its generalization properties are inherently unpredictable and unstable.
\end{abstract}

\section{Introduction}

The goal of distributional alignment is to match---as closely as possible---the output distribution of LLMs to the distribution of human values, or to some other reference distribution. This property is important especially as LLMs are increasingly being used to simulate or probe human behavior. For example, social scientists now use LLMs to simulate economic games, conduct preliminary user studies or marketing interviews, and so on, all of which requires proper representation of the humans they aim to generate insights from \cite{horton2023large}. Distributionally aligned LLMs can also be useful for the general population. For example, consider a voter in an upcoming election in California who is interested in understanding how Californians might vote on a particular ballot measure. An LLM that fully understands the distribution of political views in California may be able to forecast the outcome of the ballot measure \textit{a priori}, without extensive polling on this particular issue. Being able to reliably simulate human opinions and values could lead to more productive and less costly research, polling, and policymaking.

More broadly, distributionally aligned LLMs are designed to better align with human values or opinions, which extends beyond the political opinion domain. For example, consider an LLM that generates book reviews that skew overwhelmingly positive, as has been documented by \cite{wu2024ai}. We might want to align the LLM's output distribution to the distribution of human book reviews, for example, from Goodreads. Having more fine-grained control over the LLM's output distribution is particularly important when we have strong priors about what we want the distribution to look like, whether that means matching the human distribution or some other normative ideal such as fairness.

An obstacle in achieving distributional alignment is  generalizing to other tasks from small training data. In the political opinions example, it remains unclear if teaching a model to be aligned on a specific political issue (e.g., the environment) will necessarily make other issues more representative as well (e.g., religion). Since LLMs have strong priors on how topics relate semantically to one another, it may suffice to train an LLM on a set of representative topics in order for it to generalize, which would have significant implications on the sample size required to distributionally align LLMs. 

\noindent \textbf{Contributions.} In this work, we aim to tackle this problem through two research questions:
\begin{enumerate}[label=\textbf{R\arabic*.}, leftmargin=*]
    \item \textbf{Achieving distributional alignment.} How can we distributionally align an LLM so that it can generalize out-of-sample?
    \item \textbf{Out-of-task/topic generalization.} After achieving distributional alignment, how does the LLM's performance change on out-of-task or out-of-topic inputs?
\end{enumerate}

We demonstrate our findings on two large-scale benchmarks: a multiple-choice benchmark on political opinions, and an open-ended benchmark based on the game \textit{Family Feud.} We find that:
\begin{enumerate}[leftmargin=1em]
    \item LLMs know how to distributionally align to unseen samples of the same domain, though alignment is not perfect.
    \item Out-of-topic generalization depends on how correlated the topics are.
    \item Distributionally aligned LLMs remain coherent and useful for out-of-task questions, but their outputs change in unpredictable ways. For example, LLMs calibrated on a Family Feud dataset lose their creative abilities.
\end{enumerate}

\section{Background and Related Work}

\subsection{Pluralistic Alignment}

Pluralistic alignment is the idea that humans fundamentally have varying opinions and values, and that LLM alignment should explicitly account for this diversity. This idea departs from traditional machine learning, which typically assumes a single ground truth for any particular task. An emerging literature investigates the extent to which LLMs are pluralistically aligned in domains such as morality \cite{scherrer2023evaluating}, political opinions \cite{santurkar2023whose}, emotions \cite{he2024emotions}, and safety \cite{kirk2024personalisation}.

One instantiation of pluralistic alignment is \textit{distributional} \cite{sorensen2024roadmap}, which is the perspective we take in this paper. However, there are other notions of alignment (e.g., overton and steerable), which may be equally valuable for other settings. While there is no shortage of work auditing and assessing LLM misalignment, few papers attempt to actually align the output distribution of LLMs to a distribution of human values. One notable paper that attempts to do this is \cite{feng2024modular}, which uses community language models to simulate different parts of the population and then feeds their outputs to an aggregator LLM to better align the output distribution. To achieve distributional alignment, this paper uses prior weights on how representative each community language model is of the overall population, for example the distribution of Democrats and Republicans in the United States. However, it assumes that each community language model is a good representation of the population it aims to simulate, which is a strong and difficult-to-test assumption. Our approach, in contrast, grounds its method in actual human values data where we have fine-grained access to the distribution of values for every question.

\subsection{Homogenization in LLM Outputs}

Homogenization in generative models is the phenomenon where the output distribution becomes less diverse. In generative adversarial networks (GANs), this phenomenon is known and well documented as ``mode collapse'' \cite{kossale2022mode}. Recently, homogenization has been a topic of interest in the LLM literature. Various studies have shown that LLMs---particularly those that have been fine-tuned or aligned via reinforcement learning from human feedback (RLHF)---exhibit this mode collapse in a variety of settings. For example, \cite{santurkar2023whose} show that fine-tuned LLMs skew left-leaning in their political opinions. Other papers have also demonstrated homogenization in LLMs in the context of creative writing \cite{doshi2024generative}, coding \cite{wu2024ai}, and more. When homogeneous LLMs are used for a task where humans have diverse values, their outputs are inevitably biased toward one part of the distribution, resulting in misalignment and representational biases.

\subsection{Social Simulacra}

Social scientists are increasingly interested in the prospect of using LLMs to simulate human systems. \cite{park2022social} and \cite{park2023generative} use language models to simulate controlled environments where agents perform everyday tasks and interact with other agents in the system. \cite{horton2023large} investigates the extent to which LLMs can reliably simulate economic agents when given information about human preferences and beliefs. See \cite{gao2023survey} for a survey of relevant work in this area.

\section{Setup}

\subsection{Problem Statement}

Consider a distribution \(P\) over some space \((\mathcal{X}, \mathcal{Y})\), where \(\mathcal{X}\) is the input space (prompts) and \(\mathcal{Y}\) is the output space (answers). For every \(x \in \mathcal{X}\), we have a distribution \(P(y \mid x)\) over \(\mathcal{Y}\), which denotes the ground truth distribution of human values. We also have an LLM that outputs a distribution \(Q(y \mid x)\) over \(\mathcal{Y}\). Our goal is to minimize
\[
    \operatorname{dist}\bigl(Q(y \mid x), P(y \mid x)\bigr),
\]
where \(\operatorname{dist}\) is some distance metric such as Wasserstein distance or total variation distance.

\subsection{Datasets and Models}

\paragraph{Datasets.}
For each of the above research questions, we use two datasets:

\textbf{1. OpinionQA} is a dataset of questions and multiple-choice answers about people's political opinions taken from Pew Research Center's American Trends Panel (ATP) \cite{santurkar2023whose,pewATP}. While the ATP is an ongoing project spanning more than a decade, the OpinionQA dataset is a curated subset of the ATP that contains 1,506 question-answer pairs over various topics, including politics, religion, and culture. One convenient feature of this dataset is that each option is mapped to a position, as shown in the example below, that quantifies distance between options. This makes measuring the Wasserstein distance straightforward, since we have a notion of how far apart each option is.

    \begin{figure}[t]
        \centering
        \includegraphics[width=\linewidth]{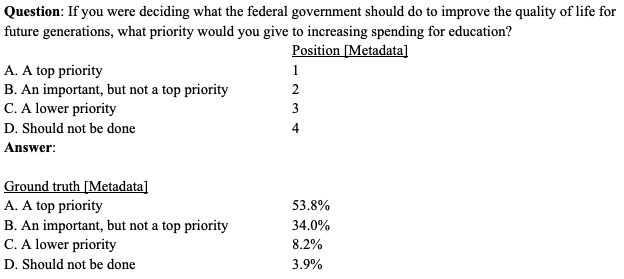}
        \caption{Example question from the OpinionQA dataset. Ground truth distribution and positions are shown for illustrative purposes and are not passed into the model.}
        \label{fig:example-question-opinionqa}
    \end{figure}

\begin{figure}[t]
        \centering
        \includegraphics[width=0.6\linewidth]{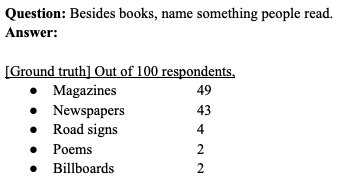}
        \caption{Example question from the Family Feud dataset. Ground truth distribution is shown for illustrative purposes and is not passed into the model.}
        \label{fig:example-question-familyfeud}
    \end{figure}

\textbf{2. ProtoQA}, which we will refer to as ``Family Feud,'' is a curated dataset of questions and answers from the game show \textit{Family Feud} \cite{boratko2020protoqa}. The basic premise is simple: a panel of 100 respondents is asked to respond to a question, making up a probability distribution of human answers. The dataset contains 8,714 question-answer pairs. Figure~\ref{fig:example-question-familyfeud} shows an example question from the dataset.

Since OpinionQA is a multiple-choice dataset with at most five choices per question, it is a much easier task than Family Feud, which requires an open-ended, short-form response. As such, the methods for achieving distributional alignment differ for the two datasets, and my analyses vary slightly.

\paragraph{Models.}
Throughout the experiments, we use the Llama-3.2 1B Instruct model \cite{meta2024llama32}.

\section{R1: Achieving Distributional Alignment}

In this section, I discuss methods for achieving distributional alignment for both datasets, as well as preliminary results. First, I introduce some notation. As before, let \((x, y) \in (\mathcal{X}, \mathcal{Y})\) be prompt-answer pairs, and let \(P(y \mid x)\) be the ground truth distribution of answers for prompt \(x\), while \(Q(y \mid x)\) is the distribution of the LLM's outputs. We disaggregate answers \(y\) into tokens indexed by position \(t \in \{0, 1, \dots, t_{\max}\}\), where \(t_{\max}\) denotes the maximum token position for an answer.

\subsection{Multiple Choice Tasks (OpinionQA)}

\textbf{Metric.} We use \textbf{Wasserstein distance} to measure the distance between the ground truth distribution \(P\) and the LLM's distribution \(Q\).

\subsubsection{Fine-tuning Method}

Multiple choice tasks are relatively easy because we need only to predict the next token, i.e., \(t = 0\). For each question \(x\), let the valid token vocabulary be
\[
    \mathcal{V}_x = \{A, B, C, \dots\},
\]
corresponding to the set of valid answers for that question. We use cross-entropy loss:
\[
    \mathcal{L}_{\mathrm{CE}}
    =
    -\sum_{x \in \mathcal{X}}
    \sum_{v \in \mathcal{V}_x}
    P(y_0 = v \mid x)
    \log Q(y_0 = v \mid x).
\]
Cross-entropy loss is a proper scoring rule, which means that the function is minimized when \(Q\) matches \(P\) in distribution.

\subsubsection{Baseline/Naive Methods}

\textbf{(a) Global temperature scaling.}
    Find the optimal temperature \(\tau^*\) that minimizes the Wasserstein distance between \(P\) and \(Q\) when we scale the logits underlying \(Q\) by \(\tau^*\), on average across the training set. Formally,
    \[
        \tau^*
        =
        \arg\min_{\tau}
        \mathbb{E}_{x \in \mathcal{X}_{\mathrm{train}}}
        \left[
            \operatorname{Wass}\bigl(P(y \mid x), Q(y \mid x; \tau)\bigr)
        \right].
    \]
    We use only this temperature to scale logits for all questions.

\textbf{(b) Question-specific temperature scaling.}
    For each question \(x\), choose a temperature \(\tau_x^*\) that minimizes the Wasserstein distance between \(P\) and \(Q\) when we scale the logits underlying \(Q\) by \(\tau_x^*\):
    \[
        \tau_x^*
        =
        \arg\min_{\tau}
        \operatorname{Wass}\bigl(P(y \mid x), Q(y \mid x; \tau)\bigr),
        \qquad
        \forall x \in \mathcal{X}.
    \]
This method may be the best post-hoc method for achieving distributional alignment because we are adjusting the logits with access to ground truth data on all questions. Note that temperature scaling cannot change the relative ordering of the logits. Therefore, when logits are misordered relative to the ground truth, no amount of temperature scaling will reach the minimum Wasserstein distance. Hence, we clip at \(\tau_x = 10\).

\subsubsection{Experiment Details}

We split the dataset into 70--15--15 train-validation-test splits. We then fine-tuned the Llama model using the train set, saved the model with the best validation loss, and evaluated on all three splits. See Table~\ref{tab:opinionqa-hyperparameters} for the list of hyperparameters used.

\subsubsection{Results}
\begin{figure}[t]
    \centering
    \includegraphics[width=\linewidth]{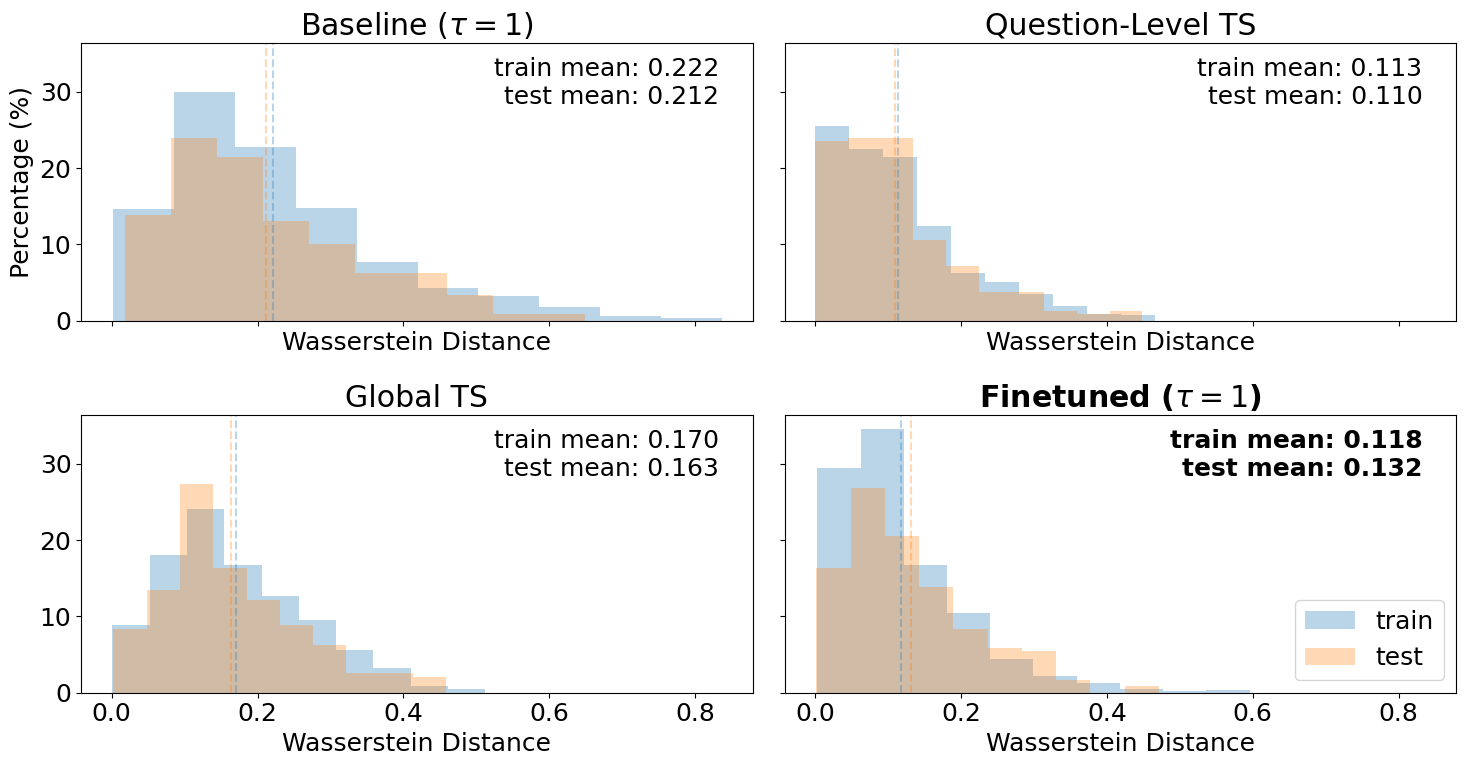}
    \caption{Distribution of Wasserstein distance for the base model (top left), base model with question-level temperature scaling (top right), base model with global temperature scaling (bottom left), and fine-tuned model (bottom right).}
    \label{fig:opinionqa-wasserstein}
\end{figure}

We only show results for the hyperparameters with the best validation loss. Figure~\ref{fig:opinionqa-wasserstein} plots the distribution of Wasserstein distance for the base model with no temperature scaling, the base model with question-level and global temperature scaling, and the fine-tuned model with no temperature scaling.

As expected, temperature scaling improves alignment relative to the base model. While the base model achieves a mean distance of \(0.22\) across the dataset, global temperature scaling reduces the mean to \(0.17\), and question-specific temperature scaling reduces the mean to \(0.11\). Note, however, that both temperature scaling methods assume access to ground truth data. In particular, question-level temperature scaling assumes ground truth on all questions, which is an unreasonable assumption, especially for the test set. Yet, the fine-tuned model achieves comparable performance to question-specific temperature scaling, with a mean distance of \(0.12\) and \(0.13\) for the train and test sets, respectively. This indicates that the fine-tuned model is able to generalize out-of-sample, even beating post-hoc temperature scaling methods.

\subsection{Open-Ended Tasks (Family Feud)}\label{sec:family_feud_r1}

\textbf{Metric.} Unlike OpinionQA, the Family Feud dataset does not have a built-in notion of distance between answers. As such, we use \textbf{total variation distance (TVD} to measure the distance between the ground truth distribution \(P\) and the LLM's distribution \(Q\).

\subsubsection{Fine-tuning Method}

Also unlike OpinionQA, answers can span multiple tokens. For example, ``umbrella'' might consist of two tokens: ``um'' and ``brella.'' As such, we introduce additional notation to account for this extra complexity. Let \(\mathcal{V}_0\) be the set of possible tokens in the first position such that
\[
    \sum_{v_0 \in \mathcal{V}_0} P(y_0 = v_0 \mid x) = 1.
\]

Let \(v_{<t}\) denote the sequence of tokens from position \(0\) to \(t-1\): $v_{<t} = (v_0, v_1, \dots, v_{t-1}).$
Finally, let \(\mathcal{V}_t(v_{<t})\) be the set of possible tokens at position \(t\), conditional on previous tokens \(v_{<t}\). In particular, we index \(\mathcal{V}_t(v_{<t})\) such that
\[
    \sum_{v_t \in \mathcal{V}_t(v_{<t})}
    P(y_t = v_t \mid x, v_{<t})
    =
    1
    \qquad
    \forall v_{<t}.
\]

Notice that the ground truth distribution \(P\) is defined with a tree structure where all conditional probabilities sum to \(1\). Figure~\ref{fig:familyfeud-tree} gives an intuitive example of how to construct this tree structure. We can now define a similar cross-entropy loss:

\begin{figure}[t]
    \centering
    \includegraphics[width=0.9\linewidth]{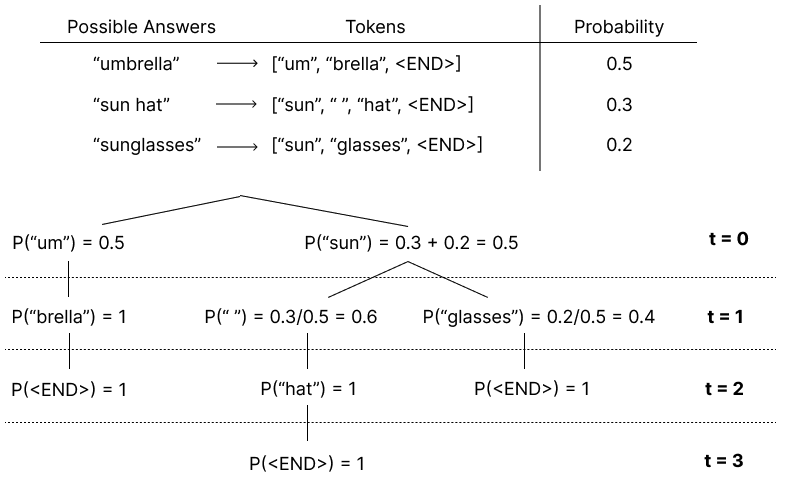}
    \caption{Example of how to construct the probability tree structure for a distribution of answers in the Family Feud dataset.}
    \label{fig:familyfeud-tree}
\end{figure}

\[
\begin{aligned}
    \mathcal{L}_{\mathrm{CE}}
    =
    &-\sum_{x \in \mathcal{X}}
    \sum_{t=0}^{t_{\max}}
    \sum_{v_{<t}}
    \sum_{v_t \in \mathcal{V}_t(v_{<t})}
    \beta_t
    P(y_t = v_t \mid x, v_{<t}) \\
    &\hspace{9em}
    {}\log Q(y_t = v_t \mid x, v_{<t}), .
\end{aligned}
\]
where
\[
    \beta_t =
    \begin{cases}
        \gamma, & \text{if } t = 0, \\
        1, & \text{if } t > 0.
    \end{cases}
\]
The purpose of \(\gamma\) is to weight the loss for the first token more heavily, since calibrating the first token impacts how the remaining tokens are generated. As before, cross-entropy loss is a proper scoring rule, which means that the function is minimized when \(Q\) matches \(P\) in distribution.

\subsubsection{Experiment Details}

Due to computational constraints, we use a 10--10--80 train-val-test split. See Table~\ref{tab:familyfeud-hyperparameters} for the list of hyperparameters used.

Since the task is open-ended, constructing a probability distribution for the generative model is not as straightforward as in OpinionQA. Ideally, we would like to query the model sufficiently many times and construct the distribution empirically. However, we need a way to cluster answers into thematically similar groups. For example, ``umbrella,'' ``Umbrella,'' and ``parasol'' are semantically similar but distinct token-wise. We propose two proxies for this distribution:

\begin{enumerate}[leftmargin=1em]
    \item \textbf{Experiment 1: Directly calculating probabilities of possible answers.}
    For each question \(x\), we take only the list of possible ground truth answers, such as ``umbrella,'' ``sun hat,'' and ``sunglasses,'' and calculate the probability of each answer, ignoring all others. Concretely, we first calculate token probabilities by taking the softmax of the logits over the entire vocabulary:
    \[
        P(\text{``um''})
        =
        \frac{\exp(z_{\text{um}})}
        {\sum_{j=1}^{V} \exp(z_j)},
    \]
    where \(z_i\) is the logit for the \(i\)-th token and \(V\) is the vocabulary size. We use a temperature of \(\tau = 1.0\). Then, we obtain the word probability via the chain rule:
    \[
        P(\text{``umbrella''})
        =
        P(\text{``um''})
        \cdot
        P(\text{``brella''} \mid \text{``um''}).
    \]
    This method is straightforward and computationally efficient, but it is slightly unfair since it assumes access to the ground truth answers. As shown in the results, temperature scaling would not have improved the baseline model because all probabilities are close to zero.

    \item \textbf{Experiment 2: Calculating probabilities from 100 generations.}
    For each question \(x\), we generate 100 words from the model using top-$k$ sampling. Then, we cluster answers semantically to generate a representative set of candidate answers. Using this set, we calculate the probability for each word using the chain rule as above, then normalize the cluster probabilities to sum to 1. See Appendix~\ref{appsec:generate_100_details} for details.
    
    This method is more computationally expensive, but it does not require access to the ground truth answers and more faithfully simulates guessing answers.
\end{enumerate}

\subsubsection{Results}

\begin{figure}[t]
    \centering
    \includegraphics[width=\linewidth]{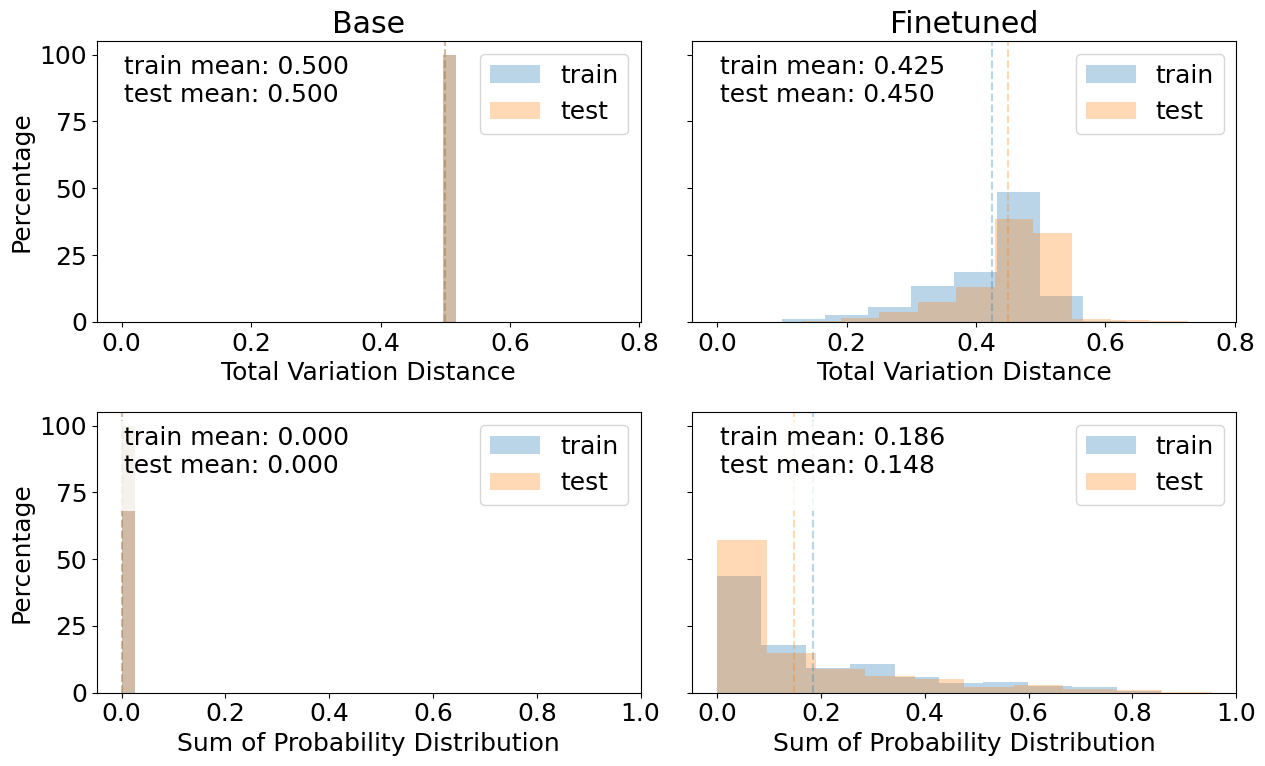}
    \caption{Distribution of total variation distance (top) and the sum of probabilities (bottom) for the fine-tuned model (right) and base model (left), when evaluated on the raw answer probabilities. All plots use temperature \(\tau = 1.0\) where relevant.}
    \label{fig:familyfeud-tvd-direct}
\end{figure}

\paragraph{Experiment 1.}
Note that since we are only calculating probabilities for valid answer tokens, our constructed \(Q\) is not strictly a probability distribution, since it does not sum to \(1\). As such, total variation distance is not the most appropriate metric. To illustrate this point, we plot both the total variation distance and the sum of the probabilities in \(Q\) in Figure~\ref{fig:familyfeud-tvd-direct}. The baseline model almost always produces probabilities that are close to \(0\), so the sum of the probabilities is close to \(0\) and total variation distance is capped at \(0.5\). Understandably, the base model is not aligned to our task.

After fine-tuning, however, we manage to shift the sum of the probability distribution away from \(0\). While it is far from \(1\), it is still a significant improvement, since this indicates that of the hundreds of thousands of possible tokens in the vocabulary, the model is concentrating logits toward valid answer tokens. This is also illustrated in the total variation distance plot of the fine-tuned model. On average, total variation distance is reduced to \(0.45\), even though the probability distribution has shifted away from \(0\).

\begin{figure}[t]
    \centering
    \includegraphics[width=0.8\linewidth]{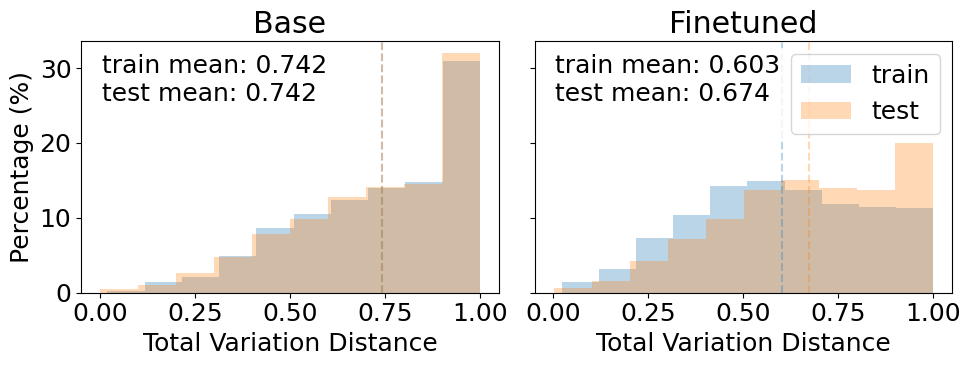}
    \caption{Distribution of total variation distance between the fine-tuned and base models, when evaluated on cluster probabilities based on 100 generated answers. Both plots use \(\tau = 1.0\).}
    \label{fig:familyfeud-tvd-cluster}
\end{figure}

\paragraph{Experiment 2.}
Unlike Experiment 1, we are now recovering a valid probability distribution \(Q\) that sums to \(1\), since we normalize the cluster probabilities. This experiment is also more representative of playing \textit{Family Feud}. Our results are displayed in Figure~\ref{fig:familyfeud-tvd-cluster}. The base model is not only poorly aligned, but also struggles with coverage: for about \(30\%\) of the questions, it fails to produce any answer that matches the ground truth distribution, as indicated by total variation distance being \(1\). After fine-tuning, this number drops to \(20\%\) for the test set and \(10\%\) for the training set.

Similarly, we see that the fine-tuned distribution is shifted to the left, with a test mean of \(0.67\) and train mean of \(0.60\), a \(7\) and \(14\) percentage point improvement over the baseline, respectively. The difference between train and test performance is likely due to slight overfitting, though this is partially mitigated by using the validation set to save the best model.

\section{R2: Out-of-topic/task Generalization}

In the previous section, we showed that fine-tuning improves distributional alignment, even on unseen samples. Here, we use the best fine-tuned model and investigate a different type of generalization: performance on out-of-topic and out-of-task samples.

\subsection{Shared Experiments}

\subsubsection{Experiment A: MMLU}

For both OpinionQA and Family Feud, we use the MMLU dataset to evaluate the model's general knowledge \cite{hendrycks2021measuring}. The MMLU dataset contains 14,042 multiple-choice questions across 57 topics spanning logic, mathematics, law, medicine, and more, each with at least 100 questions. This dataset only has one correct answer per question. As such, we convert this problem into a binary classification task. Concretely, we take the maximum of the answer-token logits, such as A, B, C, and so on, and treat this as the prediction. The prediction is either correct or incorrect, and we calculate recall to evaluate the model. We also calculate precision, but the model never has false positives, so precision is always \(1\) across the board.

Additionally, we measure the entropy of the logits over the answer tokens to measure the confidence or perplexity of the model's predictions. For both entropy and recall, we compare to baseline model performance. To summarize, we use two metrics:
\[
    \text{Pct Change Recall}
    =
    \left(
        \frac{\text{Recall}_{\mathrm{ft}} - \text{Recall}_{\mathrm{base}}}
        {\text{Recall}_{\mathrm{base}}}
    \right)
    \times 100\%,
\]
and
\[
    \Delta \text{Entropy}
    =
    \text{Entropy}_{\mathrm{ft}}
    -
    \text{Entropy}_{\mathrm{base}},
\]
where the subscripts \(\mathrm{ft}\) and \(\mathrm{base}\) refer to the fine-tuned and baseline models, respectively.

\subsubsection{Experiment B: Qualitative Analysis of Free-Form Questions}

We designed a simple set of questions to test the fine-tuned model's coherence and ability to respond to unrelated questions. The questions are as follows:
\begin{itemize}[leftmargin=1em]
    \item \textbf{Question 0:} Hello, how are you?
    \item \textbf{Question 1:} What is your favorite color?
    \item \textbf{Question 2:} What is the scientific method?
    \item \textbf{Question 3:} What are your thoughts on the death penalty?
    \item \textbf{Question 4:} What is the capital of France?
    \item \textbf{Question 5:} What do you think about universal healthcare?
    \item \textbf{Question 6:} Tell me a short story.
    \item \textbf{Question 7:} Tell me a joke.
\end{itemize}

We then generated 100 responses with temperature \(\tau = 1.0\) for each question for both the base and fine-tuned models, and analyzed the responses for general trends.

\subsection{OpinionQA}

\subsubsection{Experiment A: MMLU}

\begin{figure}[t]
    \centering
    \includegraphics[width=\linewidth]{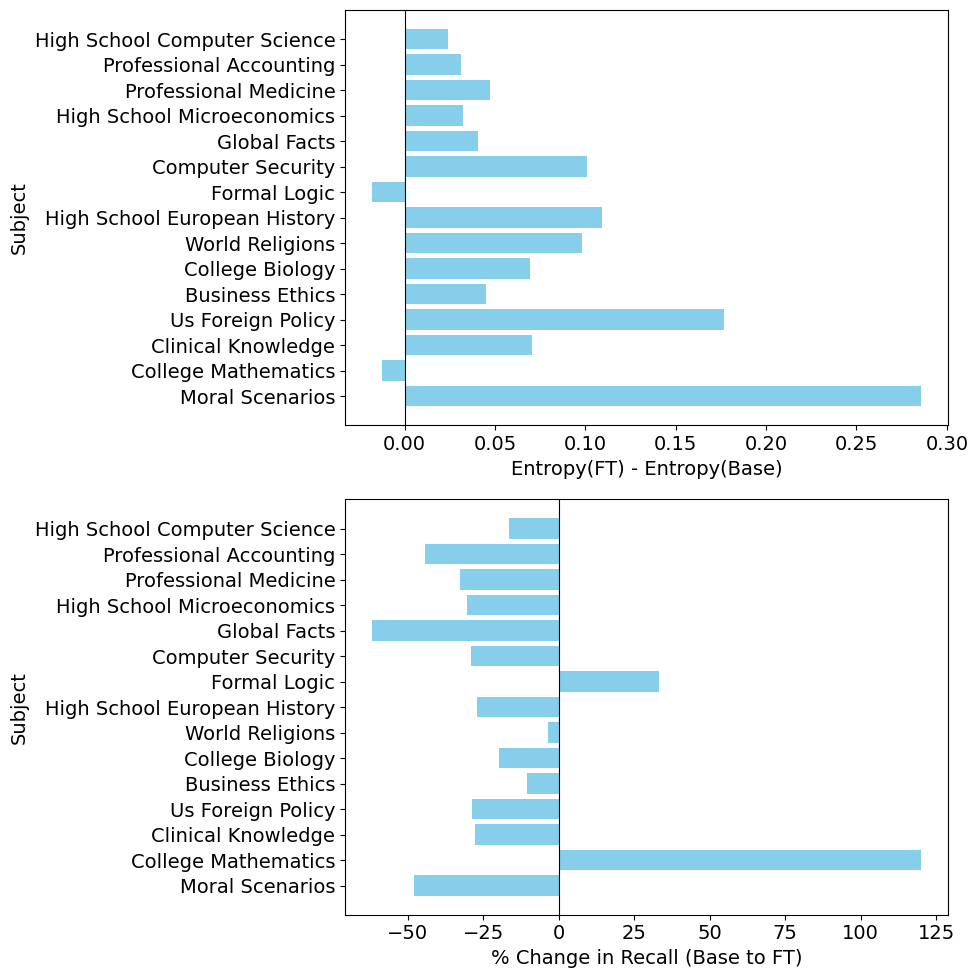}
    \caption{Change in entropy (top) and percentage change in recall (bottom) between base and OpinionQA fine-tuned models across 15 randomly chosen topics.}
    \label{fig:mmlu-opinionqa}
\end{figure}

Figure~\ref{fig:mmlu-opinionqa} displays change in entropy and recall between the base and fine-tuned models across 15 topics, chosen randomly for visual clarity. Almost all topics show a positive change in entropy and a negative change in recall, indicating that the fine-tuned model is more perplexed by the task; in other words, \textbf{fine-tuning on OpinionQA degrades the model's general knowledge.}
Our hypothesis is that both OpinionQA and MMLU are multiple-choice question-answering tasks, and explicitly fine-tuning on alphabet choices somehow makes the model more confused about alphabet choices on out-of-topic questions.

\subsubsection{Experiment B: Qualitative Analysis of Free-Form Questions}

We see no discernible difference between the base and fine-tuned models. This makes sense since we are only fine-tuning on alphabet choices, which should not break or change free-form responses drastically. See the qualitative analysis for Family Feud for more interesting results.

\subsubsection{Experiment C: Cross-Topic Generalization}

In the original fine-tuning process, we randomly split the dataset into train-validation-test splits regardless of topic. Here, we systematically bucket the dataset by topic, fine-tune on each topic, and evaluate on questions from other topics. Note that the OpinionQA dataset was too small for this experiment, so we scraped additional data from ATP. The ATP releases surveys in ``waves,'' where each wave tackles a different topic. Since these waves span more than a decade, we only obtained data from 2016--2019, as well as COVID-related questions from 2020, in order to control as much as possible for temporal effects.

Below is a list of topics and general descriptions:
\begin{itemize}
    \item \textbf{COVID.} Relating to the COVID-19 pandemic.
    \item \textbf{Environment.} Questions related to the environment, including climate change, pollution, and conservation.
    \item \textbf{Culture Surrounding Technology.} Topics include online dating, social media, and parental controls over children using technology.
    \item \textbf{Gender.} Gender identity, roles, and stereotypes.
    \item \textbf{Politics.} U.S. politics, including gun control, healthcare, and taxation, as well as international relations.
    \item \textbf{Trust.} Trust in institutions and misinformation.
    \item \textbf{Technology.} AI and large technology companies.
    \item \textbf{Religion.} Religious beliefs, practices, and how they relate to politics, family values, and other domains.
\end{itemize}

\begin{figure}[t]
    \centering
    \includegraphics[width=\linewidth]{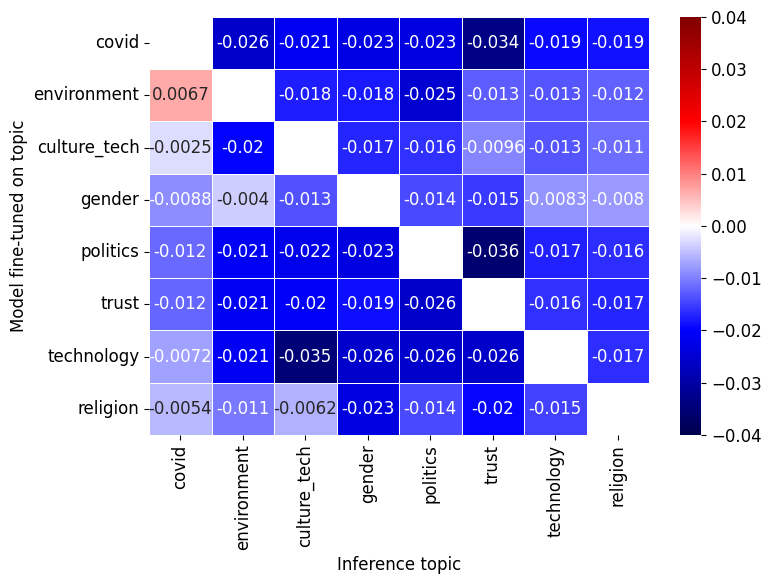}
    \caption{Heatmap of the difference in Wasserstein distance between the fine-tuned and base models, when fine-tuned on each topic (rows) and evaluated on questions from other topics (columns). Lower values indicate better alignment.}
    \label{fig:opinionqa-cross-topic-generalization}
\end{figure}

\paragraph{Results.}
Figure~\ref{fig:opinionqa-cross-topic-generalization} shows a heatmap of the difference in Wasserstein distance between the fine-tuned and base models, when fine-tuned on each topic, shown in rows. Virtually all cells except one are negative, indicating that fine-tuning on any topic improves alignment on others. This likely indicates that topics are not independent, which makes sense considering that some questions related to, for example, religion explicitly intersect with politics or gender.

Three correlations are particularly noteworthy because they lead to the largest improvements in alignment: COVID \(\rightarrow\) Trust, Politics \(\rightarrow\) Trust, and Technology \(\rightarrow\) Culture Surrounding Technology. These associations also align with our priors: the pandemic led to distrust in institutions and truthfulness; politics is often a proxy for people's trust in institutions; and people's attitudes toward AI and tech companies are likely correlated with their attitudes toward other aspects of technology.

\subsection{Family Feud}\label{sec:family_feud_r2}

\subsubsection{Experiment A: MMLU}

In contrast to OpinionQA, fine-tuning on Family Feud improves recall across the board and reduces entropy, meaning that \textbf{the model is more confident and more accurate in its predictions}. Perhaps this is because Family Feud and MMLU are different tasks but have more correlated topics. The tasks being different means that fine-tuning on short-form text may not break the model's ability to answer multiple-choice questions with one correct answer. Furthermore, we believe that Family Feud data is much more related to MMLU than OpinionQA because it contains general knowledge about people's behaviors, random objects, and random facts. This contrasts with OpinionQA data, which contains people's opinions on sensitive topics that are somewhat unrelated to general-purpose knowledge.

\begin{figure}[t]
    \centering
    \includegraphics[width=\linewidth]{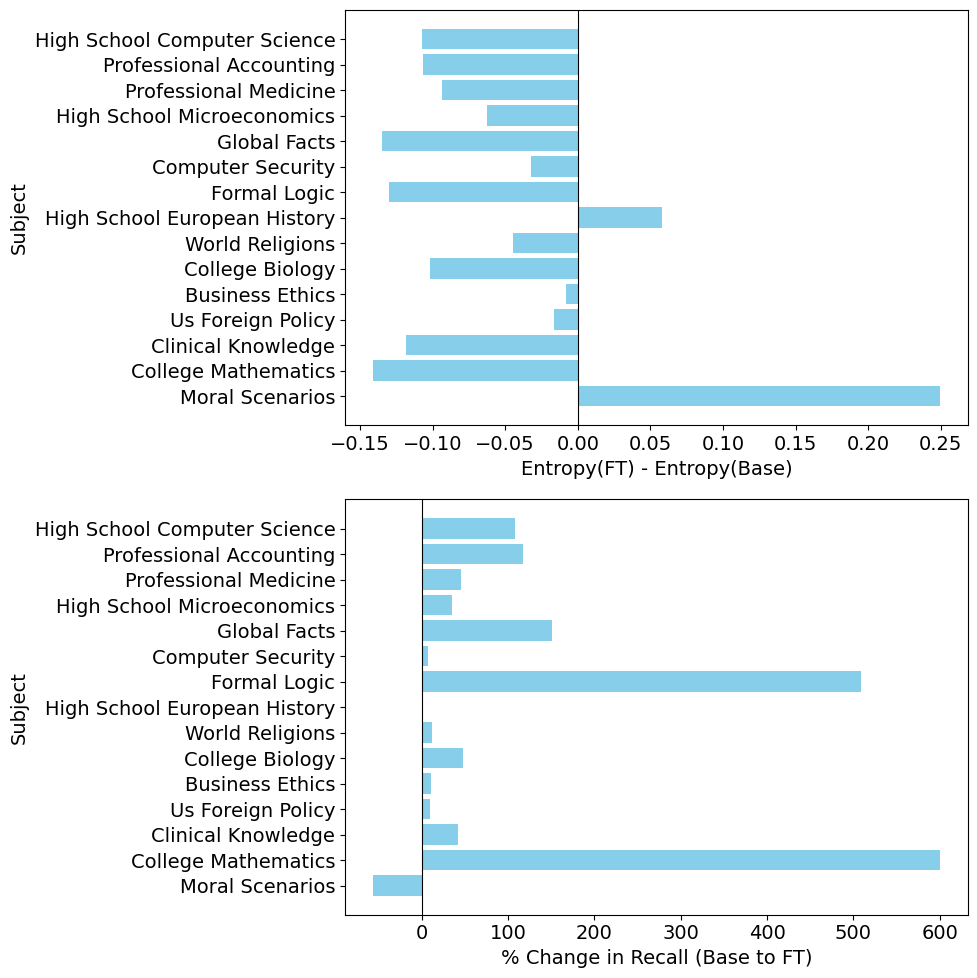}
    \caption{Change in entropy (top) and percentage change in recall (bottom) between base and Family Feud fine-tuned models across 15 randomly chosen topics.}
    \label{fig:mmlu-familyfeud}
\end{figure}

\subsubsection{Experiment B: Analysis of Free-Form Questions}

\begin{figure}[t]
    \centering
    \includegraphics[width=\linewidth]{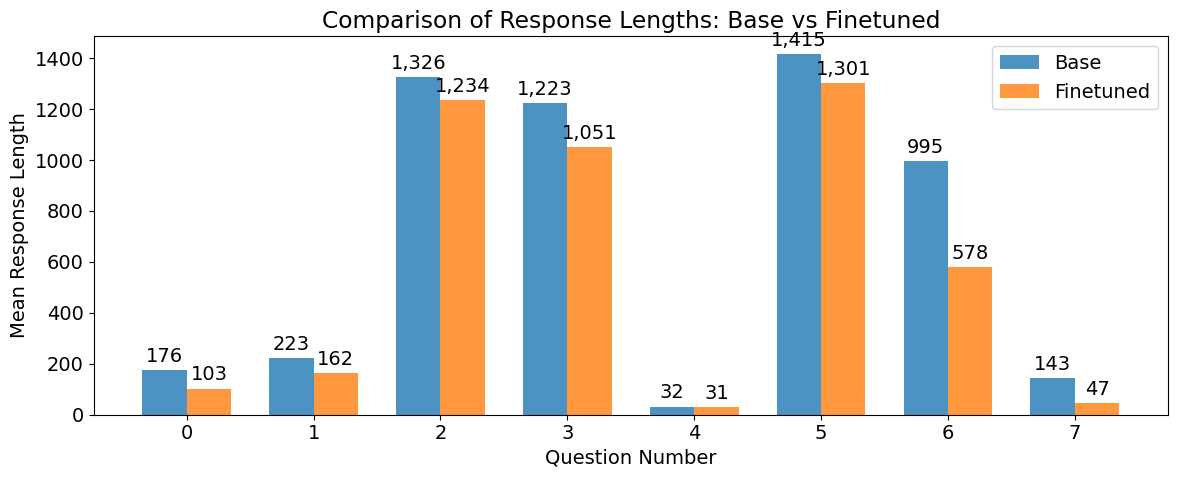}
    \caption{Average number of characters, a proxy for token length, generated for each question.}
    \label{fig:familyfeud-response-lengths}
\end{figure}

First, we show the average number of characters generated for each question in Figure~\ref{fig:familyfeud-response-lengths}, which is a proxy for token length. In contrast to the OpinionQA fine-tuned model, the Family Feud model consistently generated fewer characters on average. This is likely because the model explicitly learned to generate the \texttt{[END]} token, leading it to generate more succinct answers.

The fine-tuned model's outputs are coherent and generally do not stray too far from the base model. However, we find that \textbf{the model significantly degrades in its creative abilities}, such as writing a short story or crafting a joke. See Appendix~\ref{appsec:qualitative_extended} for detailed responses.

\section{Discussion}

\subsection{Main Takeaways}

\textbf{Distributional alignment is possible, and LLMs know how to generalize on unseen samples.} OpinionQA and, to a lesser extent, Family Feud exhibit much better alignment compared to post-hoc methods like temperature scaling on the test set.

\textbf{Out-of-topic generalization depends on how correlated the topics are.} For example, in the cross-topic generalization experiment (Figure~\ref{fig:opinionqa-cross-topic-generalization}), training on Politics or COVID increases alignment with respect to Trust in institutions and misinformation.

\textbf{Distributionally aligned LLMs remain coherent and useful for out-of-task questions, but their outputs change in unpredictable ways.} While fine-tuning on alphabet tokens in OpinionQA does not significantly change LLM outputs, training on few-token datasets such as Family Feud changes the model's ability to be empathetic and generate creative responses, such as writing short stories or telling jokes.

\subsection{Limitations and Future Work}

We conclude by outlining several limitations and how future work may address them.

\textbf{Inconclusive theories as to why OpinionQA and Family Feud perform differently on MMLU.} While we posed hypotheses as to why both fine-tuned models perform differently, it remains unclear \textit{how} to test these more rigorously. This is where mechanistic interpretability may be able to provide more satisfying explanations.

\textbf{Out-of-task generalization.}
There is no shortage of benchmark datasets that could more rigorously stress-test the fine-tuned model. However, we believe that the qualitative analysis would be a stronger method, partially because choosing benchmarks leads to a selection problem. An interesting direction for future work is to develop a systematic protocol to both generate diverse question prompts and analyze the resulting outputs automatically.

\textbf{Distributional alignment beyond the token level.} In many applications, alignment is not simply measured at the token level. For example, suppose we have a feature \(f\) that captures the political leaning or sentiment of tweets, and a distribution of values over \(f\). Operationalizing these higher level concepts remains an open question.

\bibliographystyle{named}
\bibliography{ijcai26}

\newpage
\appendix

\section{R1 Experiment Details}

\subsection{Hyperparameters}

In order to achieve distributional alignment, we tuned various hyperparameters and chose based on a held-out validation set. Tables~\ref{tab:opinionqa-hyperparameters} and~\ref{tab:familyfeud-hyperparameters} show the hyperparameters users for \textbf{OpinionQA} and \textbf{Family Feud}, respectively.

\begin{table}[t]
    \centering
    \begin{tabular}{ll}
        \toprule
        \textbf{Hyperparameter} & \textbf{Value} \\
        \midrule
        Learning rate & \(\{0.0001, 0.00005\}\) \\
        Batch size & \(4\) \\
        Number of epochs & \(50\) \\
        Gradient clipping & \(10.0\) \\
        Gradient accumulation steps & \(4\) \\
        Patience & \(8\) \\
        Scheduler & \texttt{CosineAnnealingLR} \\
        \bottomrule
    \end{tabular}
    \caption{Hyperparameters used for OpinionQA fine-tuning.}
    \label{tab:opinionqa-hyperparameters}
\end{table}

\begin{table}[t]
    \centering
    \begin{tabular}{ll}
        \toprule
        \textbf{Hyperparameter} & \textbf{Value} \\
        \midrule
        \(\gamma\) & \(\{20, 10, 5\}\) \\
        Learning rate & \(\{0.0001, 0.00005\}\) \\
        Batch size & \(8\) \\
        Number of epochs & \(50\) \\
        Gradient clipping & \(10.0\) \\
        Gradient accumulation steps & \(4\) \\
        Patience & \(8\) \\
        Scheduler & \texttt{CosineAnnealingLR} \\
        \bottomrule
    \end{tabular}
    \caption{Hyperparameters used for Family Feud fine-tuning.}
    \label{tab:familyfeud-hyperparameters}
\end{table}

\section{R2 Experiment Details}
\subsection{Details of Section~\ref{sec:family_feud_r1} (Experiment 2)}\label{appsec:generate_100_details}

In Experiment 2, we proposed calculating probabilities from 100 generations. For each question $x$, we did the following:
\begin{enumerate}
        \item Generate 100 answers from the model using top-\(k\) sampling with \(k = 6\) and temperature \(\tau = 1.0\), capping the number of tokens at 8 so that we are only considering short answers. As above, temperature is \(1\) because we are already fine-tuning the model and want to measure calibration without the need for temperature scaling. This leaves one degree of freedom, \(k\), and we choose \(k = 6\) because there are often at most six possible answers per question.

        \item Cluster the answers into semantically similar groups using agglomerative clustering with cosine distance threshold \(0.65\). Each cluster is then mapped to the answer in the ground truth distribution, if it exists, again using cosine distance with threshold \(0.65\). We use Sentence-BERT to generate embeddings of candidate answers \cite{reimers2019sentencebert}.

        \item Calculate the probability for each word via softmax and the chain rule, then add up the probabilities within each cluster.

        \item Normalize the cluster probabilities to sum to \(1\).
    \end{enumerate}

\subsection{Extended Results from Section~\ref{sec:family_feud_r2} (Experiment B)}\label{appsec:qualitative_extended}

We find that the fine-tuned model significantly degrades in its creative abilities. Below, we show representative model responses.

\begin{itemize}
    \item \textbf{Question 0: Hello, how are you?}

    While the base model is empathetic, the fine-tuned model is more functional. The base model manages to answer the question while still abstaining from claiming human emotions, but the fine-tuned model refuses to answer directly and instead deflects to another topic.

    \begin{quote}
    \texttt{Base: I'm doing well, thanks for asking. I'm a large language model, so I don't have feelings or emotions like humans do, but I'm always ready to help with any questions or topics you'd like to discuss. How about you? How's your day going so far?}
    \end{quote}

    \begin{quote}
    \texttt{Fine-tuned: I'm a large language model, I don't have emotions like humans do, but I'm functioning properly and ready to assist you. How can I help you today?}
    \end{quote}

    \item \textbf{Question 6: Tell me a short story.}

    The base model generates short stories normally. The fine-tuned model prematurely cuts off stories about one in three times. It also repeats stories about a girl in a village more often than the base model, indicating some form of homogenization.

    \item \textbf{Question 7: Tell me a joke.}

    The base model generates a joke normally, with a setup and punchline. However, the fine-tuned model consistently cuts off at the setup and fails to deliver the punchline. Interestingly, the base model mainly generates only one type of joke, while the fine-tuned model generates more variety.

    \begin{quote}
    \texttt{Base: A man walked into a library and asked the librarian, ``Do you have any books on Pavlov's dogs and Schrödinger's cat?'' The librarian replied, ``It rings a bell, but I'm not sure if it's here or not.''}
    \end{quote}

    \begin{quote}
    \texttt{Fine-tuned: Why did the scarecrow win an award?}
    \end{quote}

    \begin{quote}
    \texttt{Fine-tuned: What does a can opener say when it gets tired?}
    \end{quote}

    \begin{quote}
    \texttt{Fine-tuned: Why couldn't the bicycle stand up by itself?}
    \end{quote}
\end{itemize}
\end{document}